\documentclass[11pt]{article}

\usepackage[preprint]{latex/acl}

\usepackage{times}
\usepackage{latexsym}

\newcommand{\method}{DeltaMoE\ }
\usepackage{amsmath}
\usepackage{amssymb}
\newcommand{\minisection}[1]{\noindent{\textbf{#1}}}
\usepackage{enumitem}
\usepackage{booktabs}
\usepackage{bm}
\usepackage{array}
\usepackage{arydshln}
\newcommand{\thead}[1]{\begin{tabular}{@{}c@{}}#1\end{tabular}}
\usepackage{rotating}

\usepackage[T1]{fontenc}

\usepackage[utf8]{inputenc}

\usepackage{microtype}

\usepackage{inconsolata}

\usepackage{graphicx}

\title{A Data-Efficient Path to Multilingual LLMs: Language Expansion via Post-training PARAM$\Delta$ Integration into Upcycled MoE}

\author{
  \textbf{Hao Zhou\textsuperscript{1}}\thanks{Work done during internship at Tongyi Lab.},
  \textbf{Tianhao Li\textsuperscript{2}},
  \textbf{Zhijun Wang\textsuperscript{1}},
  \textbf{Shuaijie She\textsuperscript{1}},
  \textbf{Linjuan Wu\textsuperscript{3}},
\\
  \textbf{Hao-Ran Wei\textsuperscript{2}}\thanks{Corresponding author.},
  \textbf{Baosong Yang\textsuperscript{2}},
  \textbf{Jiajun Chen\textsuperscript{1}},
  \textbf{Shujian Huang\textsuperscript{1}}\footnotemark[2]
\\
  \textsuperscript{1}National Key Laboratory for Novel Software Technology, Nanjing University \\
  \textsuperscript{2}Tongyi Lab, Alibaba Group \textsuperscript{3}Zhejiang University
\\
  {\{zhouh,wangzj,shesj\}@smail.nju.edu.cn, \{chenjj,huangsj\}@nju.edu.cn}\\
  {\{chongsheng.lth,funan.whr,yangbaosong.ybs\}@alibaba-inc.com}\\
  {wulinjuan525@zju.edu.cn}
}

\begin{document}
\maketitle

\begin{abstract}
Expanding Large Language Models~(LLMs) to new languages is a costly endeavor, demanding extensive Continued Pre-Training~(CPT) and data-intensive alignment. While recent data-free merging techniques attempt to bypass alignment by fusing a multilingual CPT-enhanced model with its instruct counterpart, they are plagued by a critical trade-off: mitigating parameter conflicts to preserve original abilities inevitably dilutes new language acquisition, and vice-versa. To resolve this conflict, we introduce \method, which upcycles a dense model into a Mixture-of-Experts~(MoE) architecture, allocating different experts to different languages. Alignment ability is then transferred by grafting a MoE-expanded parameter delta~($\Delta_{\text{post}}$) to the CPT-enhanced base model, bypassing the complex alignment phase. Experiments demonstrate \method's superiority even against baselines with similar FLOPs or number of parameters; it improves performance on expanded languages while effectively preserving original capabilities. We further show our approach is highly applicable across different models and Post-training deltas.
\end{abstract}

\section{Introduction}
Large Language Models~(LLMs) such as LongCat-Flash~\citep{team2025longcat}, Kimi k2~\citep{team2025kimi}, Deepseek~\citep{guo2025deepseek}, have demonstrated remarkable capabilities on a variety of tasks~\citep{jiang2024survey,wang2025survey,luo2025large,wang2024factuality}. However, they remain primarily optimized for English, given that the majority of the pre-training corpus is in English. As a result, the models yield inferior results in 
non-English languages.

\begin{figure}[t]
  \centering
  \includegraphics[width=1.0\linewidth]{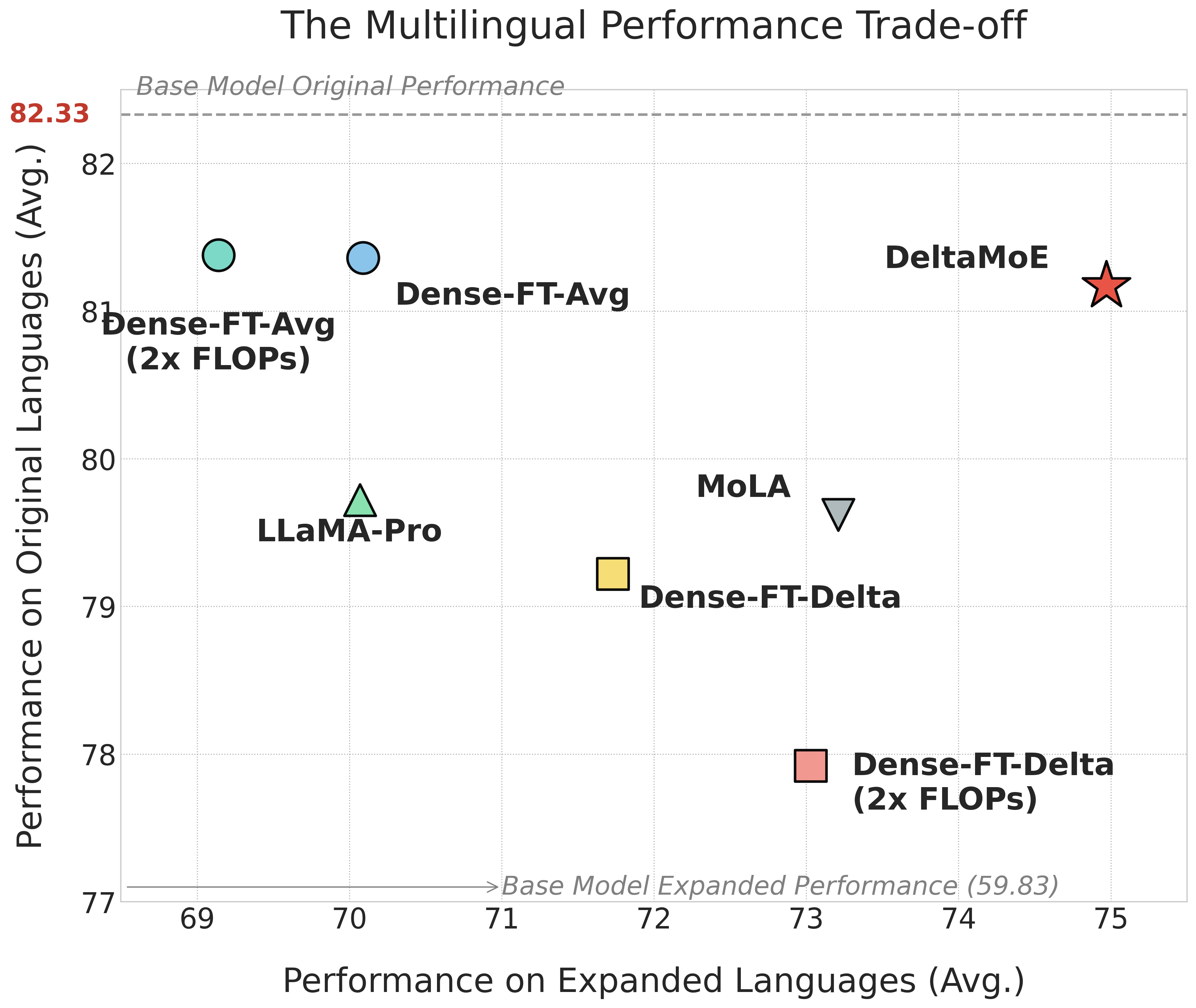}
  \caption{A visualization of the performance trade-off between expanded language capabilities (x-axis) and original language retention (y-axis). \method{} resolves this conflict better than baseline methods.}
  \label{fig:tradeoff}
\end{figure}

The standard training pipeline of expanding LLMs to other languages involves Continued Pre-Training~(CPT) followed by post-training. CPT requires massive data replay to prevent catastrophic forgetting, a costly prerequisite for the effective alignment step. The subsequent alignment stage is even more demanding, requiring not only immense computational power but also vast amounts of quality instruction data. The scale of this challenge is exemplified by the alignment of Qwen2.5~\citep{yang2025qwen3}, where each step incurs substantial costs: (1) Supervised Fine-Tuning (SFT) on millions of meticulously curated examples and (2) a complex two-stage Reinforcement Learning (RL) process involving Offline RL~\citep{rafailov2023direct} and Online RL~\citep{shao2024deepseekmath}. Together, the demands for high-quality data and large-scale computation make this pipeline prohibitively costly and difficult to achieve, highlighting the need for a more efficient alternative.

Therefore, recently, \citet{yamaguchi2024elchat, cao2025paramdelta} have explored gradient-free methods to transplant alignment capabilities from a well-aligned LLM to a continually pre-trained version of its base model. The core idea of these methods is conducting parameter merging from the post-trained model and the CPT one. However, prevalent merging techniques exist a critical trade-off between original language and expanded language. Specifically, \citet{yamaguchi2024elchat}~(similar to our \textsc{Dense-FT-Avg} baseline) uses simple linear merge~\citep{wortsman2022model} to average the weights of the post-CPT model with the original instruct model. This operation inherently discounts the CPT updates, substantially diluting the newly acquired knowledge in expanded language and thus limiting gains on expanded languages. On the other hand, delta merging~\citep{cao2025paramdelta}~(similar to our \textsc{Dense-FT-Delta} baseline) directly applies the full parameter changes from CPT, often inducing a drastic shift from the original weights, leading to catastrophic forgetting of its capabilities in the original language.

\begin{figure*}[t]
  \centering
  \includegraphics[width=1.0\linewidth]{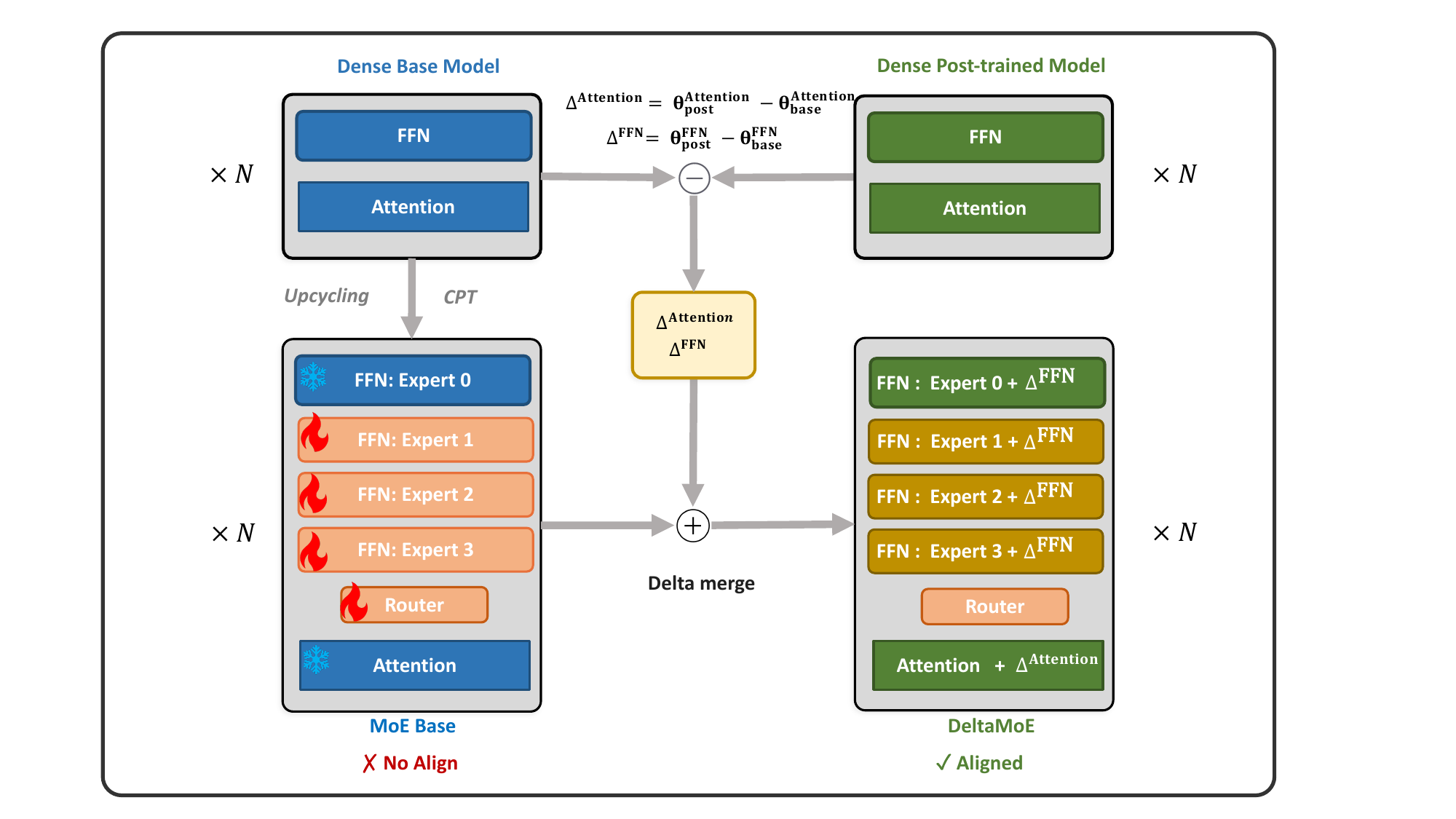}
  \caption{The two-stage \method{} pipeline: 1) CPT via sparse upcycling with a frozen expert to preserve knowledge, followed by 2) MoE model merging to transfer alignment abilities.}
  \label{fig:main_method}
\end{figure*}

To address this critical trade-off, we introduce \method, a novel approach that integrates the Mixture-of-Experts~(MoE) architecture with a refined delta merging strategy. Specifically, we first upcycle the dense base model into an MoE structure. The original parameters are frozen to serve as a dedicated repository for the original knowledge during the CPT phase. Subsequently, the delta merging principle is applied to all experts, enhancing models with alignment capabilities. This two-stage strategy, illustrated in Figure~\ref{fig:main_method}, enables \method to effectively inherit alignment capabilities on expanded languages from a well-trained open-source LLM, entirely bypassing the need for instruction data. 

Experiments show that \method outperforms strong baselines, improving average performance on expanded languages by 1.7 points and preservation capabilities by 1.5 points over other delta-based methods with comparable FLOPs or parameter counts, demonstrating our \method{} effectiveness and efficiency.

Our contributions are described as follows:
\begin{itemize}

    \item We propose a novel approach that effectively expand new languages for existing LLM and preserve original languages abilities, while acquiring alignment ability without a heavy post-training procedure. 
    \item We conduct extensive experiments demonstrating that \method{} resolves the key trade-off, substantially enhancing performance on expanded languages while mitigating catastrophic forgetting. 
    \item We confirm the robustness and generality of this approach across various base models and post-training deltas.

\end{itemize}

\section{Method}
Our proposed method in Figure~\ref{fig:main_method}, \method, is a two-stage process designed to efficiently build an expanded language-enhanced alignment model. The first stage, CPT, augments the model with new languages by selectively training only new experts and router, which preserves the original model's knowledge. The second stage, MoE Model Merging, then innovatively grafts a parameter delta from the original dense instruct model onto our sparse MoE architecture, effectively transferring alignment capabilities.

\subsection{Continued Pre-training via Sparse Upcycling}
The primary objective of this stage is to augment a pre-existing dense LLM with knowledge of new languages, while crucially preserving its original language capabilities. To achieve this, we employ a sparse upcycling strategy, transforming the dense model into an MoE Architecture. 

\minisection{MoE Architecture.}
Following the upcycling paradigm~\citep{komatsuzaki2023sparse}, we initialize $N$ experts by creating deep copies of the original feed-forward network (FFN) from the dense base model. As our base models employ SwiGLU-based FFNs~\citep{grattafiori2024llama,yang2025qwen3}, the forward pass of a single expert $E_{i}$ is defined as:
\begin{equation}
  \label{eq:moe_ffn}
  E(x)_i = (SiLU(xW^{i}_{\text {gate}} \odot xW^{i}_{\text{up}}))W^{i}_{\text{down}}
\end{equation}
where $x\in \mathbb{R}^{\text{d}}$ is the input hidden state, $W^{i}_{\text {gate}},W^{i}_{\text{up}} \in \mathbb{R}^{\text{d} \times \text{f}},W^{i}_{\text{down}} \in \mathbb{R}^{\text{f} \times \text{d}}$ are the weight matrices of the $i$-th expert. $\text{d}$ is model's hidden state dimension and $\text{f}$ is the intermediate FFN dimension.

\minisection{Top-k Gating.}
A router, parameterized by $W_{router}$ determines the contribution of each expert for a given token. The gating weights $P \in \mathbb{R}^{N}$ are computed via a softmax over the router's logits:
\begin{equation}
\label{eq:moe_prob}
p=\operatorname{softmax}\left(x W_{\text {router }}\right)
\end{equation}
We employ top-$k$ routing, which activates only the $k$ experts with the highest gating weights for each token. Let $\mathcal{T} = \operatorname{TopK}(p,k)$ be the set of indices for the $k$ selected experts. The weights for these activated experts are re-normalized:
\begin{equation}
  \label{eq:moe_nor}
  w_{i}=\left\{\begin{array}{ll}
    \frac{p_{i}}{\sum_{j \in \mathcal{T}} p_{j}} & \text { if } i \in \mathcal{T} \\
    0 & \text{otherwise}
    \end{array}\right.
\end{equation}
The final output of the MoE layer is a weighted combination of the expert outputs, combined with a residual connection:
\begin{equation}
\label{eq:moe_output}
y=\sum_{i=1}^{N} w_{i} \cdot E_{i}(x) + x
\end{equation}

\minisection{Training Objective.}
To preserve the original language knowledge, we freeze all parameters of the dense base model, including the $0$-th expert, which serves as a knowledge anchor. Consequently, only the newly added expansion experts and the router are updated during CPT~\citep{zhou2025moe}. The primary objective is the Next Token Prediction~(NTP) loss, computed as follows:
\begin{equation}
\label{eq:moe_NTP}
\mathcal{L}_{NTP} = -\frac{1}{|\mathcal{D}|} \sum_{\mathbf{y} \in \mathcal{D}} \frac{1}{|\mathbf{y}|} \sum_{t=1}^{|\mathbf{y}|} 
\log P\left(y_t \mid y_{<t}; \theta_{\text{tr}}\right)
\end{equation}
Here, $D$ is the training corpus and $\theta_{tr}$ represents the set of trainable parameters, which includes only the expansion experts and the router: $\theta_{\mathrm{tr}} = \{\theta_{\mathrm{exp}}^{(k)}\}_{k=1}^{N} \cup \{\theta_{\mathrm{router}}\}$ 

To mitigate the issue of unbalanced expert allocation, we use $\mathcal{L}_{LB}$:
\begin{equation}
\label{eq:moe_LB}
\mathcal{L}_{L B}=N \cdot \sum_{i=1}^{N} f_{i} \cdot P_{i}
\end{equation}
where $N$ is the total number of experts, $f_{i}$ is the fraction of tokens in a batch dispatched to expert $i$, $P_{i}$ is the average router probability for expert $i$ across the batch, and $\alpha$ is a scalar hyperparameter.
The final training objective is the combination of these two losses:
\begin{equation}
\label{eq:final_loss}
\mathcal{L} = \mathcal{L}_{NTP} + \alpha\mathcal{L}_{LB}
\end{equation}
where $\alpha$ is the hyper-parameter.

\subsection{MoE Model Merging}
Upon completion of the CPT stage, we obtain an expanded language-enhanced MoE base model, denoted as $M_{\text {MoE-base}}$. While this model possesses broad multilingual knowledge, it lacks the alignment capability. To instill these abilities without costly post-training, we propose a novel MoE Delta Merging strategy. This strategy adapts the delta parameterization concept~\citep{cao2025paramdelta} to our unique MoE architecture.
The core idea is to compute a delta weight $\Delta_{\text {post }}=\theta_{\text {post-trained }}-\theta_{\text {base }}$ representing the knowledge gained during the post-training of a public, dense LLM. We then graft this $\Delta_{\text {post}}$ onto our $M_{\text {MoE-base}}$ to create the final alignment model, $M_{\text {MoE post-trained}}$

Given that our $M_{\text {MoE-base}}$ is a sparse MoE model while the delta is derived from dense models, a direct application is not feasible. We therefore devise a component-wise merging strategy:

\minisection{Shared Parameters:}
For parameters that are common to both the dense and MoE architectures (e.g: attention and embedding block), we directly apply the corresponding delta weight. Let $\theta^{\text{MoE}}_{\text{shared}}$ be such a parameter in our model, and $\Delta^{\text{MoE}}_{\text{shared}}$ be the corresponding delta from the dense models. The merged parameter is:
\begin{equation}
\hat{\theta}^{\text{MoE}}_{\text{shared}} = \theta^{\text{MoE}}_{\text{shared}} + \Delta^{\text{shared}}_{\text{post}}
\end{equation}

\minisection{Expert Parameters:} 
For the expert FFN layers, which do not have a direct counterpart in the dense model's delta, we leverage the fact that they were initialized from the dense model's FFN. We compute a single FFN-specific delta, $\Delta^{\text {FFN}}_{\text {post }}=\theta^{\text{FFN}}_{\text {post }}-\theta^{\text{FFN}}_{\text {base }}$. This $\Delta^{\text {FFN}}_{\text {post }}$ is then applied uniformly to all expert parameters within our MoE base model. For the $i$-th expert's weights $\left\{W_{\text {gate }}^{i}, W_{\text {up }}^{i}, W_{\text {down }}^{i}\right\}$ the merging process is:
\begin{equation}
\begin{aligned}
\label{eq:moe_merge_experts}
\hat{W}^{i}_{\text{gate}} &= W^{i}_{\text{gate}} + \Delta^{\text{gate}}_{\text{post}} \\
\hat{W}^{i}_{\text{up}} &= W^{i}_{\text{up}} + \Delta^{\text{up}}_{\text{post}} \\
\hat{W}^{i}_{\text{down}} &= W^{i}_{\text{down}} + \Delta^{\text{down}}_{\text{post}} \\
\end{aligned}
\end{equation}

In essence, we treat all experts as having inherited the same foundational structure, and thus they should all benefit from the same post-training update derived from the dense FFN. This strategy elegantly resolves the architectural mismatch and allows the post-training knowledge to be broadcast across all experts.

\begin{table*}[t!]
\centering
\small 
\setlength{\tabcolsep}{4.5pt} 
\begin{tabular}{l rr cccccc c c}
\toprule
& \multicolumn{2}{c}{Parameters (B)} & \multicolumn{8}{c}{Zero-shot Performance} \\
\cmidrule(lr){2-3} \cmidrule(lr){4-11}
& & & & & & & \multicolumn{2}{c}{Flores-200} & \\
\cmidrule(lr){8-9}
{\bf Model} & {\bf Active } & {\bf Total} & {\bf MGSM} & {\bf MIFEVAL} & {\bf BELEBELE } & {\bf M\_MMLU} & {\bf En-XX} & {\bf XX-En} & {\bf Avg.} \\
\midrule

\multicolumn{11}{l}{\it Part 1: Performance on Expanded Languages~(hu, sr, bn)} \\
\midrule
Qwen2.5-7B-Instruct & 7.6 & 7.6 & 60.40  & 59.27  & 71.89  & 47.29  & 40.31  & 79.84  & 59.83 \\
\cmidrule(lr){1-11}

\multicolumn{11}{l}{\it Baselines w/ Same CPT Data} \\
Dense-FT-Avg       & 7.6 & 7.6 & \underline{66.13}  & 58.82  & 76.85  & 52.01  & 76.71  & 90.01  & 70.09 \\
Dense-FT-Delta     & 7.6 & 7.6 & 61.87  & 60.80  & 80.00  & 54.48  & 84.14  & 89.11  & 71.73 \\
\cmidrule(lr){1-11}

\multicolumn{11}{l}{\it Baselines w/ Matched FLOPs (2x CPT Data)} \\
Dense-FT-Avg-2FLOPs    & 7.6 & 7.6 & 62.13  & 56.74  & 76.48  & 50.66  & 78.45  & \underline{90.38}  & 69.14 \\
Dense-FT-Delta-2FLOPs  & 7.6 & 7.6 & 64.67  & \underline{62.49}  & \underline{81.44}  & 54.64  & 85.18  & 89.73  & 73.03 \\
\cmidrule(lr){1-11}

\multicolumn{11}{l}{\it Baselines w/ Matched Parameters} \\
LLaMA-Pro         & 10.8 & 10.8 & 60.80 &  58.72 &  81.70 &  \underline{55.01} &  82.56 &  86.82 &  70.94 \\
MoLA              & 13.2  & 21.7 & 65.47 & 61.44 & 80.74 & 54.52 & \underline{87.36} & 89.74 & \underline{73.21} \\
\cmidrule(lr){1-11}

\method           & 13.3 & 24.7 & \textbf{67.60}  & \textbf{64.95}  & \textbf{82.22}  & \textbf{56.11}  & \textbf{88.13}  & \textbf{90.82}  & \textbf{74.97} \\

\midrule
\multicolumn{11}{l}{\it Part 2: Performance on Original Languages~(en, zh, es, fr)} \\
\midrule
Qwen2.5-7B-Instruct & 7.6 & 7.6 & 72.50  & 77.27  & 90.47  & 69.50  & 89.94  & 94.33  & 82.33 \\
\cmidrule(lr){1-11}

\multicolumn{11}{l}{\it Baselines w/ Same CPT Data} \\
Dense-FT-Avg       & 7.6 & 7.6 & 71.20  & 72.94  & \textbf{89.56}  & \textbf{68.68}  & \textbf{90.87}  & \textbf{94.95}  & \underline{81.36} \\
Dense-FT-Delta     & 7.6 & 7.6 & 68.20  & 73.08  & 88.78  & 65.69  & 86.66  & 92.89  & 79.22 \\
\cmidrule(lr){1-11}

\multicolumn{11}{l}{\it Baselines w/ Matched FLOPs (2x CPT Data)} \\
Dense-FT-Avg-2FLOPs    & 7.6 & 7.6 & \textbf{74.30}  & 72.58  & 88.17  & \underline{67.96}  & \underline{90.41}  & \underline{94.86}  & \textbf{81.38} \\
Dense-FT-Delta-2FLOPs  & 7.6 & 7.6 & 67.00  & 72.28  & 87.47  & 63.67  & 84.55  & 92.57  & 77.92 \\
\cmidrule(lr){1-11}

\multicolumn{11}{l}{\it Baselines w/ Matched Parameters} \\
LLaMA-Pro         & 10.8 & 10.8 & 69.30 &  68.13 &  88.06 &  64.86 &  89.04 &  92.68 &  78.68 \\
MoLA              & 13.2 & 21.7 & 68.80  & \textbf{73.77}  & 88.50  & 64.43  & 89.09  & 93.15  & 79.62 \\
\cmidrule(lr){1-11}

\method        & 13.3 & 24.7 & \underline{73.00}  & \underline{73.47}  & \underline{89.42}  & 67.24  & 89.88  & 94.04  & 81.17 \\

\bottomrule
\end{tabular}
\caption{
Main results on expanded and original languages. ``Total'' denotes the total number of model parameters, while ``Active'' refers to the activated parameters during inference. The best results are in \textbf{bold}, and second-best are \underline{underlined}. Detailed performance breakdowns are available in Appendix~\ref{app:detailed_results}.
}
\label{tab:main}
\end{table*}

\section{Experiment}
\subsection{Setup}
\paragraph{Models.}
Our primary experiments use the Qwen2.5-7B~\citep{yang2025qwen3} series as the backbone, and we validate the generalizability of our approach on the LLaMA-3.1-8B~\citep{grattafiori2024llama} family in Section~\ref{sec:generalization_model}. These models were selected for their strong English performance and vocabularies well-suited for multilingual CPT. For our MoE architecture, we upcycle the dense model into a 4-expert MoE \footnote{We find that 3 trainable experts provide sufficient capacity to accommodate multiple expanded languages. See Appendix~\ref{app:expert_capacity} for detailed experiments.} with a top-2 gating strategy.

\paragraph{Training Details.}

All of our experiments are implemented using the LLaMA-Factory~\citep{zheng2024llamafactory} and are optimized for large-scale training with DeepSpeed ZeRO-3~\citep{rajbhandari2020zero}. During the CPT stage, we train for 1 epoch. We set learn rate to $5e-5$ with a cosine learning rate scheduler. The global batch size is set to 512, with a maximum sequence length of 2048 tokens. All training is performed using BF16 mixed-precision. The load-balancing loss coefficient $\alpha$ is set to 0.01.
\paragraph{Datasets.}
We designate Hungarian~(Hu), Serbian~(Sr), and Bengali~(Bn) as our expanded languages, selected due to the poor performance of the LLM on them. We also include high-resource original languages: English~(En), Chinese~(Zh), Spanish~(Es), and French~(Fr).

For each of the three expanded languages, we sample 3 billion tokens of unlabeled, monolingual text data. The data for Hungarian and Bengali are sourced from the FineWeb2 dataset~\citep{penedo2025fineweb2}. As the volume of Serbian data in FineWeb2 is insufficient for our needs, we sourced the Serbian corpus from CulturaX~\citep{nguyen2023culturax}.

\paragraph{Evaluation Benchmarks.}
To comprehensively assess zero-shot multilingual capabilities, we evaluate our models on a diverse suite of benchmarks. This includes tests for mathematical reasoning~(\textsc{MGSM}; \citealp{huang2025benchmax}),instruction following ~(\textsc{MIFEval}; \citealp{huang2025benchmax}), reading comprehension (\textsc{Belebele}; \citealp{bandarkar2023belebele}), general knowledge ~(\textsc{M\_MMLU}; \citealp{alexandra_institute_2025}), and machine translation~(\textsc{FLORES-200}; \citealp{costa2022no}). Detailed descriptions of each benchmark, including specific prompting strategies and evaluation metrics, are provided in Appendix~\ref{app:benchmarks}.

\paragraph{Baselines.}
To rigorously evaluate the effectiveness of our proposed method, \method{}, we compare it against baselines grouped into three categories for fair comparison: 1) methods with identical training data, 2) a matched computational budget (FLOPs), and 3) a comparable number of parameters.

\begin{itemize}[leftmargin=*, itemsep=3pt, topsep=2pt]

    \item \textbf{\textsc{Dense-FT-Avg}}: This baseline is a variant of the method proposed by \citet{yamaguchi2024elchat}. It performs CPT on the public instruct model, then restores alignment capabilities by linearly averaging its weights with the original instruct model.

    \item \textbf{\textsc{Dense-FT-Delta}}: Adopting the approach from \citet{cao2025paramdelta}, this baseline starts with the base dense model, performs CPT, and then adds the pre-computed alignment delta, $\Delta_{\text{instruct}}$, to instill alignment abilities.

    \item \textbf{\textsc{Dense-FT-Avg-2FLOPs} \& \textsc{Dense-FT-Delta-2FLOPs}}: To match the training FLOPs of our top-2 MoE architecture, these dense baselines are trained on twice the amount of CPT data (18B tokens total). Note that our top-2 MoE introduces approximately $1.8\times$ the training and inference FLOPs of a dense model; however, since the MoE optimizer updates all parameters, we conservatively use $2\times$ the CPT data to ensure a fair comparison.

    \item \textbf{\textsc{LLaMA-Pro}}: This baseline implements the block expansion strategy from \citet{wu2024llama}. We select the strongest configuration in Appendix~\ref{app:llamapro_ablation}. After CPT, the alignment delta is applied only to the original dense parameters.

    \item \textbf{\textsc{MoLA}}: We implement MoLA~\citep{gao2024higher} by adding LoRA~\citep{hu2022lora} experts (rank=1120) to each linear layer. The number of experts per layer increases with model depth.\footnote{Specifically, we divide the model's layers into four blocks and assign 2, 4, 6, and 8 LoRA experts to the linear layers within each respective block, from shallow to deep.} Similar to LLaMA-Pro, the alignment delta is added only to the original dense model weights.

\end{itemize}
A detailed summary of the CPT hyperparameters for all baselines is provided in Appendix~\ref{app:hyperparams}.

\subsection{Main Results}

As presented in Table~\ref{tab:main}, \method{} demonstrates two key advantages. Firstly, it establishes state-of-the-art~(SOTA) performance on the expanded languages. Secondly, it exhibits strong performance preservation on the original languages, offering a far more effective resolution to the inherent trade-off between expanding and retaining knowledge.

\paragraph{Dense Merging Reveals a Performance Trade-off.}
The dense model baselines trained on the same data reveal a stark performance trade-off. While Dense-FT-Delta surpasses Dense-FT-Avg by 1.64 points in expanded languages, it incurs a significant drop of 2.14 points in original languages. This trade-off originates from their underlying mechanics. The linear interpolation of Dense-FT-Avg ($\theta_{\text{merged}} = \frac{1}{2}(\theta_{\text{post-trained}} + \theta_{\text{post-trained cpt}})$) dilutes the newly learned knowledge, limiting its gains. Conversely, the parameter shift in Dense-FT-Delta leads to severe catastrophic forgetting of the original abilities. In contrast, \method{} circumvents this issue. \method{}'s MoE architecture decouples knowledge acquisition from retention, thus achieving a superior performance balance.
\paragraph{Superiority under Matched FLOPs.}
Even when dense baselines are given twice the CPT data to match \method training FLOPs, \method{}'s superiority holds. Although Dense-FT-Delta-2FLOPs improves on expanded languages by about 3 points to its 1x-data counterpart, it still lags behind \method{} by a significant margin of 1.94 points. More critically, its catastrophic forgetting worsens by 1.3 points compared to its 1x-data counterpart. This suggests that simply scaling data is an ineffective strategy for delta-based methods. Meanwhile, Dense-Avg-2FLOPs performs even worse on expanded languages than its 1x-data counterpart, confirming that its averaging mechanism severely dilutes new knowledge regardless of data volume.
\paragraph{Architectural Advantage of \method.}
Among parameter-expansion architectures, \method{} proves superior. The vertical expansion of LLaMA-Pro is suboptimal, as it lags behind \method{} by a substantial 4 points on expanded languages. While the stronger MoLA baseline is more competitive, it still lags behind our method by 1.76 points on expanded and 1.55 points on original languages. We attribute this performance gap to a fundamental architectural incompatibility: the dense alignment delta cannot be applied to MoLA's LoRA experts. In contrast, our method's architectural design enables the direct application of the alignment delta to our experts.

\section{Analysis and Ablation}

\begin{figure}[t]
  \centering
  \includegraphics[width=1.0\linewidth]{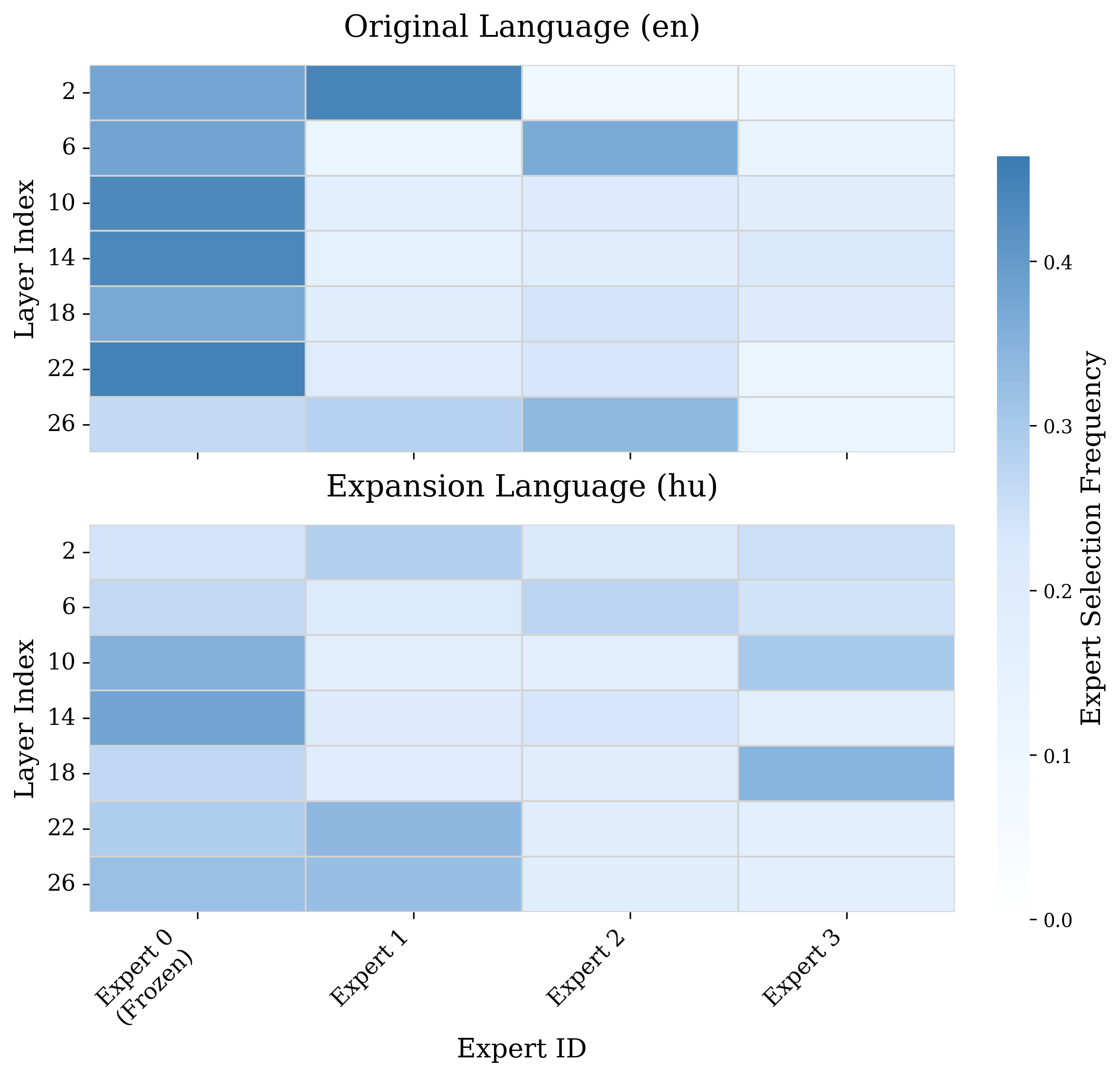}
  \caption{Average expert selection frequency across layer for English and Hungarian inputs in the ifeval benchmark.}
  \label{fig:routing_heatmap}
\end{figure}

\subsection{Analysis of Expert Routing Allocation}
As shown in Figure~\ref{fig:routing_heatmap}, the router demonstrates clear language-based specialization. For English, it routes nearly all tokens to the frozen $0$-th expert, preserving original-language capabilities. Conversely, expert selection for Hungarian shows a dynamic division of labor: trainable expansion experts are active in the upper layers, while the frozen 0-th expert handles middle-layer computations. This suggests the model performs core reasoning in English~\citep{wendler2024llamas}.

\subsection{Effective and Efficient Knowledge Retention}
\label{sec:retention_analysis}

\begin{figure}[t]
  \centering
  \includegraphics[width=1.0\linewidth]{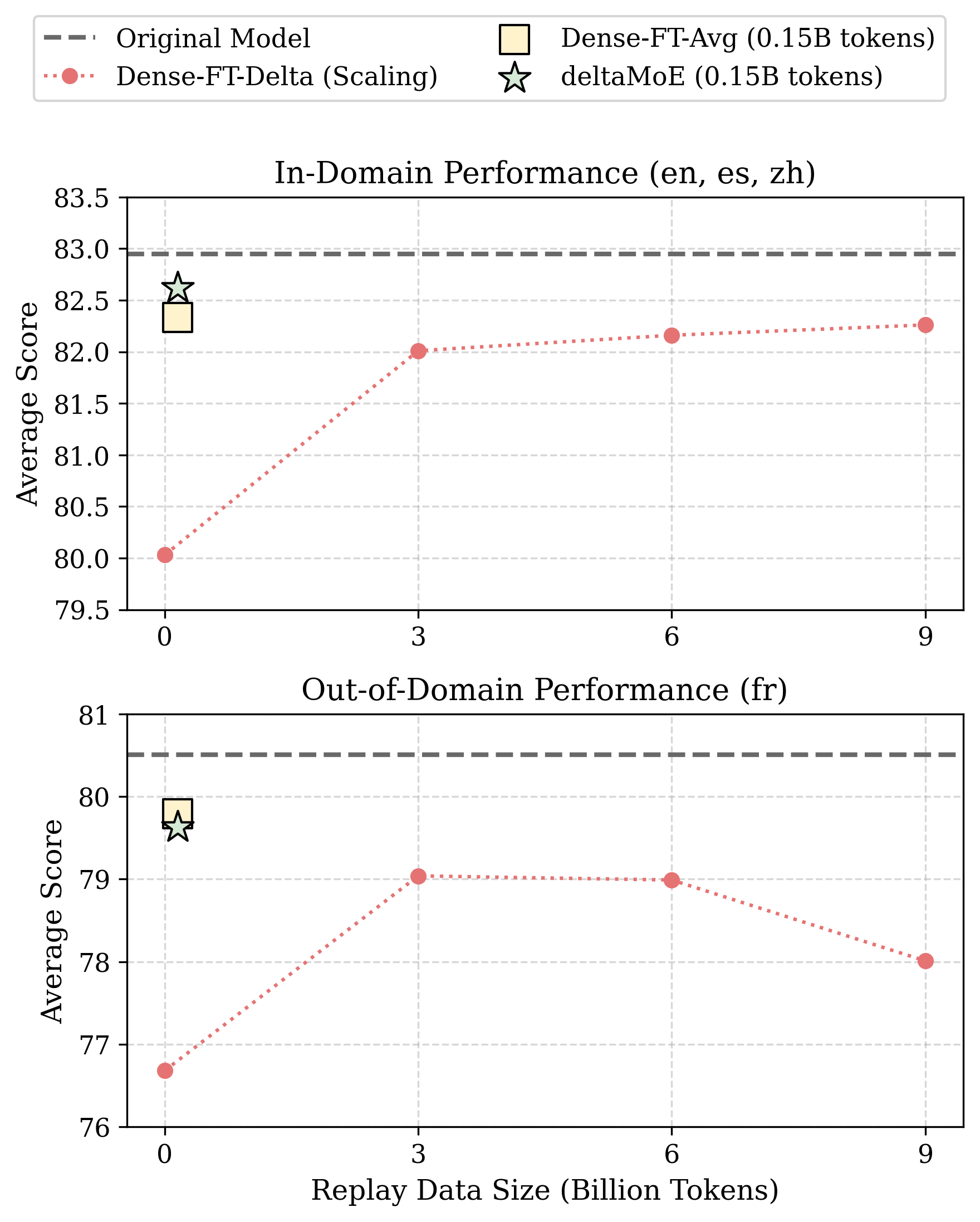}
  \caption{Knowledge retention performance with data replay. The dashed line indicates the original model's performance.}
  \label{fig:replay_scaling_law}
\end{figure}

The retention scores in our main results (Table~\ref{tab:main}) motivate this targeted analysis. While \method{} slightly trails Dense-FT-Avg in retention, it significantly outperforms Dense-FT-Delta. We therefore investigate two aspects: 1) whether minimal data replay can close the retention gap with Dense-FT-Avg, and 2) the architectural efficiency of \method{} over its dense delta counterpart.

To investigate this, we use a minimal 0.15B token in-domain replay budget (en, es, zh) and hold out French as an out-of-domain~(OOD) test\footnote{Sourced from DCLM~\citep{li2024datacomp}~(English), FineWeb2~\citep{penedo2025fineweb2}~(Spanish), and SkyPile-150B~\citep{wei2023skywork}~(Chinese), containing 50k documents per language.}. This data is applied via two distinct strategies: standard data mixing during CPT for dense models, versus a targeted post-CPT router-tuning phase\citep{zhou2025moe} for \method~(Appendix~\ref{app:router_tuning}).

Figure~\ref{fig:replay_scaling_law} shows our approach is both highly effective and efficient. With this minimal replay budget, the router-tuning enables \method to surpass Dense-FT-Avg's retention on in-domain languages. For the OOD language, it achieves highly competitive performance, closely matching the retention of Dense-FT-Avg. Simultaneously, our method demonstrates superior architectural efficiency, outperforming a Dense-FT-Delta model even when the latter is scaled with 60$\times$ more replay data in both in-domain and OOD languages. This confirms that our anchored MoE architecture, combined with router tuning, provides a more efficient solution for knowledge preservation.

\subsection{Ablation on MoE Merging Strategy}
\label{sec:ablation_moe_merge}

In order to validate the necessity of our delta merging strategy when combined with the MoE architecture, we compare against a baseline, MoE-CPT-Avg, which starts from the dense instruct model for CPT, but then applies linear averaging by merging all post-CPT MoE parameters with their counterparts in the original instruct model.

The results in Table~\ref{tab:moe_merge_ablation} show that the averaging strategy fails to resolve the fundamental performance trade-off, even within the MoE architecture. \textsc{MoE-CPT-Avg} nearly perfectly preserves original capabilities, but at the cost of severely reduced gains in expanded languages, lagging more than 4 points behind \method. This confirms that averaging inherently dilutes the crucial knowledge gained during CPT. In contrast, our delta-based approach shows a powerful synergy: the frozen expert anchors original knowledge, while delta merging transfers alignment abilities without dilution, yielding a far superior overall balance.

\begin{table}[t!]
\centering
\small
\setlength{\tabcolsep}{5pt}
\begin{tabular}{l c c c}
\toprule
\textbf{Merging Strategy} & \thead{Expanded} & \thead{Original} & \thead{ Avg.} \\
\midrule
Qwen2.5-7B-Instruct & 59.83 & 82.33 & 71.08 \\
\hdashline
MoE-CPT-Avg & 70.93 & \textbf{81.35} & 76.14 \\
\method{} (Ours) & \textbf{74.97} & 81.17 & \textbf{78.07} \\
\bottomrule
\end{tabular}
\caption{Performance comparison of different merging strategies on the MoE architecture.}
\label{tab:moe_merge_ablation}
\end{table}

\section{Generalization Analysis}
\label{sec:generalization}
To demonstrate the generalizability of \method, we conduct experiments across two dimensions: using an alignment delta ($\Delta_{\text{instruct}}$) from a different source, and applying our framework to the LLaMA-3.1-8B model family. These results confirm that our approach is not limited to a specific post-training pipeline or base architecture.

\subsection{Generalization to a Different Alignment Delta}

To verify that \method is not dependent on a specific post-training pipeline, we expanded our generalization analysis. We created a new delta Tulu-Delta by performing SFT on the Qwen2.5-7B base model using the Tulu3 dataset mixture~\citep{lambert2024tulu}\footnote{\url{https://huggingface.co/datasets/allenai/tulu-3-sft-mixture}}. This delta was then applied to \method and a comprehensive set of baselines.

The results, summarized in Table~\ref{tab:tulu_delta}, reaffirm the robustness of our framework. \method{} once again delivers the best overall performance, achieving SOTA performance on the expanded languages and competitive results that rival the top-performing baseline on the original languages. This confirms that our framework is agnostic to the source of the alignment delta and robustly transfers alignment abilities more effectively than alternative merging strategies.

\begin{table}[t!]
\centering
\small
\setlength{\tabcolsep}{5pt}
\begin{tabular}{l c c}
\toprule
\textbf{Model} & \thead{Expanded} & \thead{Original} \\
\midrule
\multicolumn{3}{l}{\textit{Base Model: Qwen2.5-7B, Delta: Tulu3}} \\
\midrule
Qwen-7B + Tulu-Delta & 53.85 & 79.90 \\
\cmidrule(lr){1-3}

\multicolumn{3}{l}{\textit{Baselines w/ Same CPT Data}} \\
Dense-FT-Avg       & 61.55 & \textbf{76.46} \\
Dense-FT-Delta     & 62.25 & 72.60 \\
\cmidrule(lr){1-3}

\multicolumn{3}{l}{\textit{Baselines w/ Matched FLOPs (2x CPT Data)}} \\
Dense-FT-Avg-2FLOPs    & 58.32 & 75.27 \\
Dense-FT-Delta-2FLOPs  & 59.64 & 68.92 \\
\cmidrule(lr){1-3}

\multicolumn{3}{l}{\textit{Baselines w/ Matched Parameters}} \\
LLaMA-Pro         & 64.97 & 75.49 \\
MoLA              & 64.25 & 74.59 \\
\cmidrule(lr){1-3}

\method{} & \textbf{65.77} & 76.14 \\
\bottomrule
\end{tabular}
\caption{
Generalization results using a delta derived from the Tulu3 SFT dataset. For LLaMA-Pro, we use its best variant in Appendix~\ref{app:llamapro_tulu_ablation}.
}
\label{tab:tulu_delta}
\end{table}

\subsection{Generalization on different model}
\label{sec:generalization_model}
\begin{table}[t]
\centering
\small
\begin{tabular}{l c c}
\toprule
\textbf{Model} & \thead{Expanded} & \thead{Original} \\
\midrule
\multicolumn{3}{l}{\textit{Base Model: LLaMA-3.1-8B}} \\
\midrule
LLaMA-3.1-8B-Instruct & 64.91 & 80.37 \\
\midrule
\multicolumn{3}{l}{\textit{Baselines w/ Same CPT Data}} \\
Dense-FT-Avg   & 66.39 & 76.54 \\
Dense-FT-Delta & 66.88 & 74.37 \\
\midrule
\multicolumn{3}{l}{\textit{Baselines w/ Matched FLOPs}} \\
Dense-FT-Avg-2FLOPs   & 65.81 & 75.78 \\
Dense-FT-Delta-2FLOPs & 64.38 & 71.62 \\
\midrule
\multicolumn{3}{l}{\textit{Baseline w/ Matched Parameters}} \\
LLaMA-Pro & 58.92 & 65.78 \\ 
MoLA & 68.39 & 74.21 \\
\midrule
\method{} & \textbf{69.37} & \textbf{77.32} \\
\bottomrule
\end{tabular}
\caption{
Generalization results on the LLaMA-3.1-8B model family. 
For LLaMA-Pro, we use its strongest variant as determined by Appendix~\ref{app:llamapro_llama3_ablation}}
\label{tab:llama_model}
\end{table}

To further assess the generality of our approach across model architecture, we replicate the entire experimental pipeline on a different, widely-used family of models LLaMA-3.1-8B~\citep{grattafiori2024llama}.

As shown in Table~\ref{tab:llama_model}, the findings are consistent with our main results. \method{} delivers the most favorable trade-off, achieving SOTA performance on both expanded and original languages. This result strongly indicates that \method{} serves as a general and robust strategy for enhancing performance on expanded languages while preserving strong capabilities in the original ones across various foundational models.

\section{Related Work}

\subsection{Mixture of Experts}

The MoE architecture enables efficient scaling of LLMs by activating only a subset of parameters per token~\citep{du2022glam, lepikhin2021gshard, zoph2022st}. This paradigm allows for models with massive parameter counts while maintaining a fixed inference budget. 
Recent advancements have further refined MoE through techniques like shared and fine-grained experts~\citep{guo2025deepseek}, zero-expert~\citep {jin2025moe} which dynamiclly control the activated parameters, as well as shortcut-connected mechanisms that optimize inference speed~\citep{cai2025shortcutconnected}.

The MoE architecture is increasingly being adopted for multilingual CPT. Early approaches in this area, such as MoE-LPR~\citep{zhou2025moe}, established a two-stage training process to balance performance across original and expansion languages. More recent methods, including DMoE~\citep{li2025group} and LayerMoE~\citep{zhang2025less}, have refined this concept by dynamically allocating experts based on linguistic similarity. However, these methods focus solely on expanding the base model, while our work presents a more holistic solution that integrates the subsequent transfer of alignment abilities.

\subsection{Model Merging}

Model merging is a data-free method for combining capabilities from multiple specialized models\citep{yu2024language,yadav2023ties}. Commonly, it is used to resolve task conflicts in post-training, such as merging specialized skills~\citep{yadav2023ties, ma2025led,wu2025unlocking,dang2024aya}.

In the context of CPT, merging has been repurposed to transfer alignment capabilities from a public model to a continually trained multilingual base model. This application, however, introduces a critical trade-off: simple averaging dilutes newly learned knowledge~\citep{yamaguchi2024elchat}, while delta-based methods can cause catastrophic forgetting of original abilities~\citep{cao2025paramdelta}. Our work introduces a novel merging strategy tailored for MoE models that directly resolves this trade-off.

\section{Conclusion}
This paper presented \method{}, a framework that resolves trade-off by first creating new experts in an MoE architecture while freezing all original parameters during CPT, then grafting an alignment delta ($\Delta_{\text{post}}$). Experiments confirm \method{} significantly enhances expansion language performance while mitigating catastrophic forgetting, proving effective across diverse models and alignment deltas. Ultimately, \method{} offers a practical and scalable pathway for extending the multilingual alignment of existing LLMs.

\section*{Limitations}
While \method{} effectively resolves the core trade-off between acquiring new languages and retaining original capabilities, this work has two primary limitations. First, the evaluated languages and benchmarks, while substantial, are insufficient to fully represent the global linguistic diversity and task spectrum. Second, the MoE architecture introduces non-trivial computational overhead in both training and inference compared to dense models.

\section*{Acknowledgments}
We would like to thank the anonymous reviewers for their insightful comments. Shujian Huang and Hao-Ran Wei are the co-corresponding authors. This work is supported by National Science Foundation of China (No. 62376116), research project of Nanjing University-China Mobile Joint Institute (NJ20250038), the Fundamental Research Funds for the Central Universities (No. 2024300507).

\bibliography{latex/custom.bib}

\clearpage
\appendix

\section{Detailed Benchmark Results}
\label{app:detailed_results}

This section provides a comprehensive breakdown of the zero-shot performance for all models on each benchmark. Tables \ref{tab:mgsm_detailed} through \ref{tab:flores_xx_en_detailed} detail the scores.

\begin{table*}[t!]
\centering
\small
\begin{tabular}{l cccc c ccc c}
\toprule
& \multicolumn{4}{c}{\textbf{Original Languages}} & & \multicolumn{3}{c}{\textbf{Expanded Languages}} & \\
\cmidrule(lr){2-5} \cmidrule(lr){7-9}
\textbf{Model} & \textbf{en} & \textbf{es} & \textbf{zh} & \textbf{fr} & \textbf{Avg.} & \textbf{hu} & \textbf{bn} & \textbf{sr} & \textbf{Avg.} \\
\midrule
\multicolumn{10}{l}{\textbf{MGSM}} \\
\midrule
Qwen2.5-7B-Instruct & 80.80 & 72.00 & 73.60 & 63.60 & 72.50 & 61.20 & 58.40 & 61.60 & 60.40 \\
\hdashline
\multicolumn{10}{l}{\it Baselines w/ Same CPT Data} \\
Dense-FT-Avg       & 80.40 &  64.40 &  75.60 &  64.40 &  71.20 &  64.40 &  64.40 &  69.60 &  66.13 \\
Dense-FT-Delta     & 78.00 &  67.20 &  73.60 &  54.00 &  68.2 &  56.80 &  63.20 &  65.60 &  61.87 \\
\hdashline
\multicolumn{10}{l}{\it Baselines w/ Matched FLOPs (2x CPT Data)} \\
Dense-FT-Avg-2FLOPs    & 80.00 & 70.80 & 78.00 & 68.40 & 74.30 & 56.80 & 62.00 & 67.60 & 62.13 \\
Dense-FT-Delta-2FLOPs  & 76.00 & 61.60 & 72.40 & 58.00 & 67.00 & 62.00 & 61.20 & 70.80 & 64.67 \\
\hdashline
\multicolumn{10}{l}{\it Baselines w/ Matched Parameters} \\
LLaMA-Pro         & 77.20 &  72.00 &  69.20 &  58.80 &  69.30 &  60.40 &  53.60 &  68.40 &  60.80 \\
MoLA              & 79.60 &  65.20 &  70.80 &  59.60 &  68.80 &  63.20 &  66.00 &  67.20 &  65.47 \\
\hdashline
\method{}           & 82.40 &  71.60 &  74.40 &  63.60 &  73.00 &  65.20 &  65.60 &  72.00 &  67.60 \\
\bottomrule
\end{tabular}
\caption{Detailed per-language results on the MGSM benchmark.}
\label{tab:mgsm_detailed}
\end{table*}

\begin{table*}[t!]
\centering
\small
\begin{tabular}{l cccc c ccc c}
\toprule
& \multicolumn{4}{c}{\textbf{Original Languages}} & & \multicolumn{3}{c}{\textbf{Expanded Languages}} & \\
\cmidrule(lr){2-5} \cmidrule(lr){7-9}
\textbf{Model} & \textbf{en} & \textbf{es} & \textbf{zh} & \textbf{fr} & \textbf{Avg.} & \textbf{hu} & \textbf{bn} & \textbf{sr} & \textbf{Avg.} \\
\midrule
\multicolumn{10}{l}{\textbf{MIFEval}} \\
\midrule
Qwen2.5-7B-Instruct & 79.40 & 77.46 & 74.72 & 77.50 & 77.27 & 58.91 & 57.78 & 61.12 & 59.27 \\
\hdashline
\multicolumn{10}{l}{\it Baselines w/ Same CPT Data} \\
Dense-FT-Avg       & 77.76 & 73.22 & 67.40 & 73.37 & 72.94 & 61.48 & 51.45 & 63.53 & 58.82 \\
Dense-FT-Delta     & 77.87 & 69.93 & 69.89 & 74.62 & 73.08 & 60.98 & 53.01 & 68.40 & 60.80 \\
\hdashline
\multicolumn{10}{l}{\it Baselines w/ Matched FLOPs (2x CPT Data)} \\
Dense-FT-Avg-2FLOPs    & 78.25 & 72.53 & 67.05 & 72.49 & 72.58 & 55.47 & 52.80 & 61.95 & 56.74 \\
Dense-FT-Delta-2FLOPs  & 77.13 & 69.34 & 69.89 & 72.77 & 72.28 & 64.67 & 56.87 & 65.94 & 62.49 \\
\hdashline
\multicolumn{10}{l}{\it Baselines w/ Matched Parameters} \\
LLaMA-Pro         & 71.73 &  66.62 &  64.64 &  69.53 &  68.13 &  60.20 &  50.40 &  65.58 &  58.72 \\
MoLA              & 78.01 & 72.79 & 71.69 & 72.59 & 73.77 & 62.98 & 52.85 & 68.50 & 61.44 \\
\hdashline
\method{}           & 76.37 & 73.84 & 69.93 & 73.72 & 73.47 & 68.54 & 55.90 & 70.40 & 64.95 \\
\bottomrule
\end{tabular}
\caption{Detailed per-language results on the MIFEval benchmark.}
\label{tab:mifeval_detailed}
\end{table*}

\begin{table*}[t!]
\centering
\small
\begin{tabular}{l cccc c ccc c}
\toprule
& \multicolumn{4}{c}{\textbf{Original Languages}} & & \multicolumn{3}{c}{\textbf{Expanded Languages}} & \\
\cmidrule(lr){2-5} \cmidrule(lr){7-9}
\textbf{Model} & \textbf{en} & \textbf{es} & \textbf{zh} & \textbf{fr} & \textbf{Avg.} & \textbf{hu} & \textbf{bn} & \textbf{sr} & \textbf{Avg.} \\
\midrule
\multicolumn{10}{l}{\textbf{Belebele}} \\
\midrule
Qwen2.5-7B-Instruct & 93.11 & 89.44 & 89.00 & 90.33 & 90.47 & 70.33 & 67.67 & 77.67 & 71.89 \\
\hdashline
\multicolumn{10}{l}{\it Baselines w/ Same CPT Data} \\
Dense-FT-Avg       & 91.78 & 88.00 & 88.44 & 90.00 & 89.56 & 76.33 & 72.44 & 81.78 & 76.85 \\
Dense-FT-Delta     & 91.67 & 87.22 & 88.67 & 87.56 & 88.78 & 84.22 & 71.22 & 84.56 & 80.00 \\
\hdashline
\multicolumn{10}{l}{\it Baselines w/ Matched FLOPs (2x CPT Data)} \\
Dense-FT-Avg-2FLOPs    & 90.89 & 86.56 & 86.78 & 88.44 & 88.17 & 74.11 & 72.67 & 82.67 & 76.48 \\
Dense-FT-Delta-2FLOPs  & 92.00 & 84.89 & 87.11 & 85.89 & 87.47 & 83.89 & 75.33 & 85.11 & 81.44 \\
\hdashline
\multicolumn{10}{l}{\it Baselines w/ Matched Parameters} \\
LLaMA-Pro         & 91.11 &  85.78 &  87.89 &  87.44 &  88.06 &  85.00 &  74.78 &  85.33 &  81.70 \\
MoLA              & 92.33 & 85.89 & 87.78 & 88.00 & 88.50 & 83.89 & 72.89 & 85.44 & 80.74 \\
\hdashline
\method{}           & 93.44 & 88.33 & 88.44 & 87.44 & 89.42 & 85.00 & 74.78 & 86.89 & 82.22 \\
\bottomrule
\end{tabular}
\caption{Detailed per-language results on the BELEBELE benchmark.}
\label{tab:belebele_detailed}
\end{table*}

\begin{table*}[t!]
\centering
\small
\begin{tabular}{l cccc c ccc c}
\toprule
& \multicolumn{4}{c}{\textbf{Original Languages}} & & \multicolumn{3}{c}{\textbf{Expanded Languages}} & \\
\cmidrule(lr){2-5} \cmidrule(lr){7-9}
\textbf{Model} & \textbf{en} & \textbf{es} & \textbf{zh} & \textbf{fr} & \textbf{Avg.} & \textbf{hu} & \textbf{bn} & \textbf{sr} & \textbf{Avg.} \\
\midrule
\multicolumn{10}{l}{\textbf{M\_MMLU}} \\
\midrule
Qwen2.5-7B-Instruct & 73.80 & 68.37 & 68.11 & 67.71 & 69.50 & 49.00 & 39.82 & 53.03 & 47.29 \\
\hdashline
\multicolumn{10}{l}{\it Baselines w/ Same CPT Data} \\
Dense-FT-Avg       & 73.34 & 68.75 & 66.44 & 66.18 & 68.68 & 56.00 & 41.54 & 58.50 & 52.01 \\
Dense-FT-Delta     & 71.61 & 63.71 & 64.24 & 63.20 & 65.69 & 58.00 & 44.07 & 61.38 & 54.48 \\
\hdashline
\multicolumn{10}{l}{\it Baselines w/ Matched FLOPs (2x CPT Data)} \\
Dense-FT-Avg-2FLOPs    & 73.12 & 66.12 & 66.36 & 66.26 & 67.96 & 51.15 & 42.11 & 58.73 & 50.66 \\
Dense-FT-Delta-2FLOPs  & 68.45 & 62.28 & 60.98 & 62.97 & 63.67 & 57.46 & 45.38 & 61.08 & 54.64 \\
\hdashline
\multicolumn{10}{l}{\it Baselines w/ Matched Parameters} \\
LLaMA-Pro         & 70.41 &  63.41 &  62.80 &  62.82 &  64.86 &  59.69 &  44.56 &  60.77 &  55.01 \\
MoLA              & 70.18 & 61.16 & 63.79 & 62.59 & 64.43 & 57.46 & 44.73 & 61.38 & 54.52 \\
\hdashline
\method{}           & 73.04 & 64.91 & 66.29 & 64.73 & 67.24 & 59.31 & 46.28 & 62.75 & 56.11 \\
\bottomrule
\end{tabular}
\caption{Detailed per-language results on the M\_MMLU benchmark.}
\label{tab:m_mmlu_detailed}
\end{table*}

\begin{table*}[t!]
\centering
\small
\begin{tabular}{l ccc c ccc c}
\toprule
& \multicolumn{3}{c}{\textbf{Original Languages}} & & \multicolumn{3}{c}{\textbf{Expanded Languages}} & \\
\cmidrule(lr){2-4} \cmidrule(lr){6-8}
\textbf{Model} & \textbf{en-es} & \textbf{en-zh} & \textbf{en-fr} & \textbf{Avg.} & \textbf{en-hu} & \textbf{en-bn} & \textbf{en-sr} & \textbf{Avg.} \\
\midrule
\multicolumn{9}{l}{\textbf{Flores-200 (En-XX)}} \\
\midrule
Qwen2.5-7B-Instruct & 91.93 & 88.89 & 89.00 & 89.94 & 34.04 & 40.44 & 46.44 & 40.31 \\
\hdashline
\multicolumn{9}{l}{\it Baselines w/ Same CPT Data} \\
Dense-FT-Avg       & 92.68 & 89.70 & 90.22 & 90.87 & 79.45 & 70.66 & 80.02 & 76.71 \\
Dense-FT-Delta     & 90.08 & 83.00 & 86.89 & 86.66 & 88.31 & 79.00 & 85.10 & 84.14 \\
\hdashline
\multicolumn{9}{l}{\it Baselines w/ Matched FLOPs (2x CPT Data)} \\
Dense-FT-Avg-2FLOPs    & 92.54 & 89.04 & 89.65 & 90.41 & 80.14 & 73.27 & 81.95 & 78.45 \\
Dense-FT-Delta-2FLOPs  & 88.53 & 80.05 & 85.05 & 84.55 & 89.89 & 80.72 & 84.92 & 85.18 \\
\hdashline
\multicolumn{9}{l}{\it Baselines w/ Matched Parameters} \\
LLaMA-Pro  & 91.53 &  87.49 &  88.12 &  89.04 & 91.01 &  82.11 &  74.57 &  82.56 \\
MoLA              & 91.10 & 88.20 & 87.98 & 89.09 & 90.56 & 82.54 & 88.97 & 87.36 \\
\hdashline
\method{}           & 92.22 & 88.49 & 88.92 & 89.88 & 91.95 & 84.21 & 88.24 & 88.13 \\
\bottomrule
\end{tabular}
\caption{Detailed per-language results for Flores-200 (En-XX) translation.}
\label{tab:flores_en_xx_detailed}
\end{table*}

\begin{table*}[t!]
\centering
\small
\begin{tabular}{l ccc c ccc c}
\toprule
& \multicolumn{3}{c}{\textbf{Original Languages}} & & \multicolumn{3}{c}{\textbf{Expanded Languages}} & \\
\cmidrule(lr){2-4} \cmidrule(lr){6-8}
\textbf{Model} & \textbf{es-en} & \textbf{zh-en} & \textbf{fr-en} & \textbf{Avg.} & \textbf{hu-en} & \textbf{bn-en} & \textbf{sr-en} & \textbf{Avg.} \\
\midrule
\multicolumn{9}{l}{\textbf{Flores-200 (XX-En)}} \\
\midrule
Qwen2.5-7B-Instruct & 94.09 & 93.99 & 94.91 & 94.33 & 77.72 & 77.17 & 84.62 & 79.84 \\
\hdashline
\multicolumn{9}{l}{\it Baselines w/ Same CPT Data} \\
Dense-FT-Avg       & 94.14 & 95.74 & 94.98 & 94.95 & 91.44 & 86.25 & 92.35 & 90.01 \\
Dense-FT-Delta     & 92.78 & 92.06 & 93.83 & 92.89 & 90.92 & 84.65 & 91.76 & 89.11 \\
\hdashline
\multicolumn{9}{l}{\it Baselines w/ Matched FLOPs (2x CPT Data)} \\
Dense-FT-Avg-2FLOPs    & 94.09 & 95.47 & 95.02 & 94.86 & 92.02 & 86.41 & 92.72 & 90.38 \\
Dense-FT-Delta-2FLOPs  & 92.87 & 91.29 & 93.54 & 92.57 & 91.64 & 85.82 & 91.75 & 89.73 \\
\hdashline
\multicolumn{9}{l}{\it Baselines w/ Matched Parameters} \\
LLaMA-Pro         & 92.05 &  93.47 &  92.53 &  92.68 &  88.91 &  84.13 &  87.42 &  86.82 \\
MoLA              & 92.99 & 92.63 & 93.81 & 93.15 & 91.83 & 85.31 & 92.08 & 89.74 \\
\hdashline
\method{}           & 93.64 & 94.05 & 94.43 & 94.04 & 92.53 & 87.25 & 92.69 & 90.82 \\
\bottomrule
\end{tabular}
\caption{Detailed per-language results for Flores-200 (XX-En) translation.}
\label{tab:flores_xx_en_detailed}
\end{table*}

\section{Baselines CPT Hyperparameter Settings}
\label{app:hyperparams}

Table~\ref{tab:cpt_hyperparams} provides a detailed summary of the key hyperparameters used during the CPT stage for all baseline models presented in the main experiments.

\begin{table*}[t]
\centering
\small
\setlength{\tabcolsep}{6pt} 
\begin{tabular}{l l c c}
\toprule
\textbf{Model / Group} & \thead{Learning \\ Rate} & \thead{Global Batch \\ Size} & \thead{CPT Data \\ (Tokens)} \\
\midrule
\multicolumn{4}{l}{\textit{Dense Merging Baselines}} \\
\quad Dense-FT (Avg \& Delta) & 2e-5 & 512 & 9B \\
\quad Dense-FT-2FLOPs (Avg \& Delta) & 2e-5 & 512 & 18B \\
\cmidrule(l){1-4}
\multicolumn{4}{l}{\textit{Parameter-Expansion Baselines}} \\
\quad LLaMA-Pro & 2e-4 & 512 & 9B \\
\quad MoLA & 5e-5 & 512 & 9B \\
\cmidrule(l){1-4}
\end{tabular}
\caption{Hyperparameter settings for the CPT stage. All baselines were trained for 1 epoch using the AdamW optimizer and a cosine learning rate scheduler.}
\label{tab:cpt_hyperparams}
\end{table*}

\section{Evaluation Benchmark and Prompting Details}
\label{app:benchmarks}

We provide a detailed description of the benchmarks and prompting strategies used to evaluate our models in a zero-shot setting.

\subsection{Benchmark Descriptions}

\begin{itemize}[leftmargin=*, itemsep=2pt, topsep=2pt]
    \item \textbf{\textsc{MGSM}}~\citep{huang2025benchmax}: A multilingual benchmark for grade-school mathematical reasoning. We adopt the standard zero-shot chain-of-thought prompting strategy from the original paper and report accuracy.

    \item \textbf{\textsc{MIFEval}}~\citep{huang2025benchmax}: A benchmark designed to test a model's adherence to complex and nuanced instructions in a multilingual context. We use the prompts and evaluation scripts provided by the authors. Following standard practice~\citep{grattafiori2024llama}, we report the overall score, which averages four sub-metrics (prompt-strict, prompt-loose, instruction-strict, and instruction-loose).

    \item \textbf{\textsc{FLORES-200}}~\citep{costa2022no}: A large-scale benchmark for machine translation. We use a custom prompt format to ensure the model directly outputs the translated text. The template is as follows in Figure \ref{fig:prompt_mt}. Performance is measured using the reference-free XCOMET-XXL metric~\citep{guerreiro2024xcomet}\footnote{\url{https://huggingface.co/Unbabel/XCOMET-XXL}} for both English-to-target (En-XX) and target-to-English (XX-En) directions.

    \item \textbf{\textsc{Belebele}}~\citep{bandarkar2023belebele}: A massively multilingual reading comprehension dataset. We report zero-shot accuracy using the chain-of-thought prompt detailed in Section~\ref{app:mcq_prompts}.

    \item \textbf{\textsc{M\_MMLU}}~\citep{alexandra_institute_2025}: A multilingual version of the MMLU benchmark for general knowledge. We report zero-shot accuracy using the chain-of-thought prompt detailed in Section~\ref{app:mcq_prompts}. To maintain representativeness while reducing computational overhead, we evaluate on a stratified subset created by sampling 10\% of questions from each subject category.
\end{itemize}

\subsection{Multiple-Choice Question Prompting and Extraction}
\label{app:mcq_prompts}

For the multiple-choice question (MCQ) benchmarks (\textsc{M\_MMLU} and \textsc{Belebele}), we employ a unified chain-of-thought prompting strategy to encourage step-by-step reasoning in zero-shot setting.

\begin{figure}
  \centering
  \includegraphics[width=1.0\linewidth]{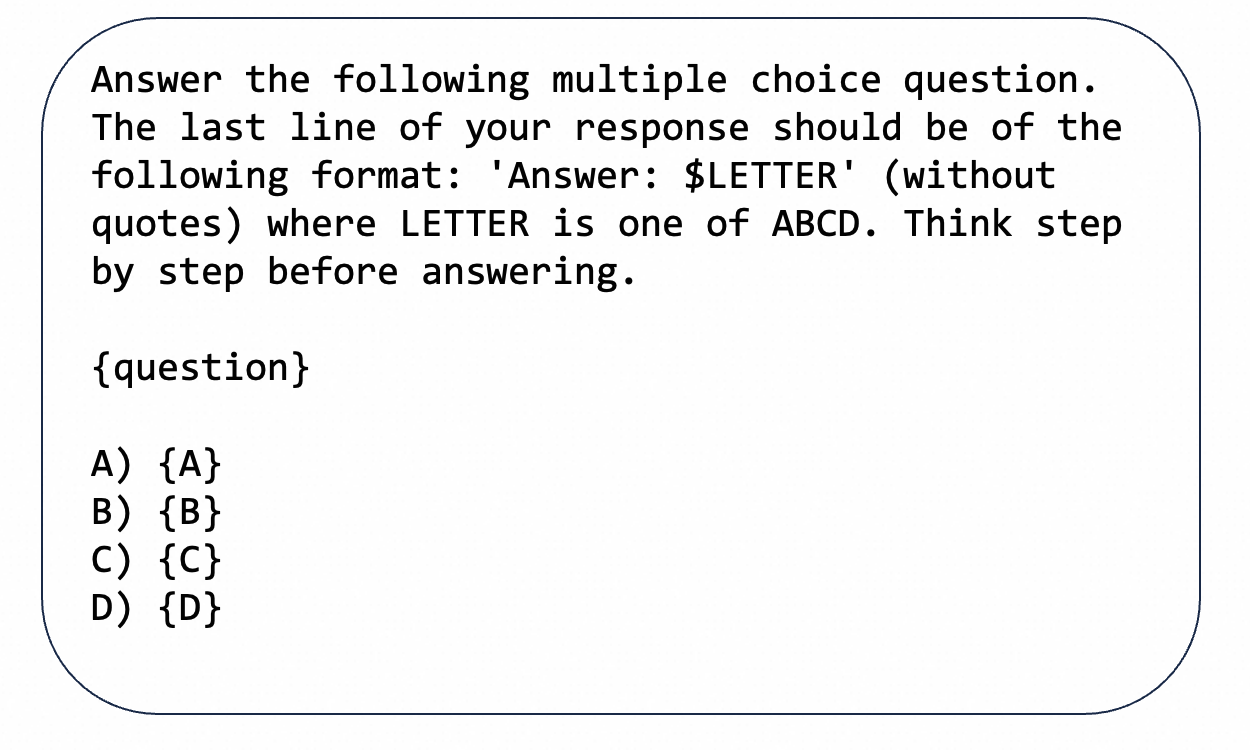}
  \caption{The prompt for mmlu evaluation}
  \label{fig:prompt}
\end{figure}

\minisection{Prompt Translation and Structure.}
We adapted the English prompt template from the OpenAI simple-evals repository\footnote{\url{https://github.com/openai/simple-evals}}. To create multilingual versions, the English template was translated into each target language using the DeepSeek-V3~\citep{guo2025deepseek} model. An example of the English MMLU prompt is shown in Figure \ref{fig:prompt}:

\begin{figure}
  \centering
  \includegraphics[width=1.0\linewidth]{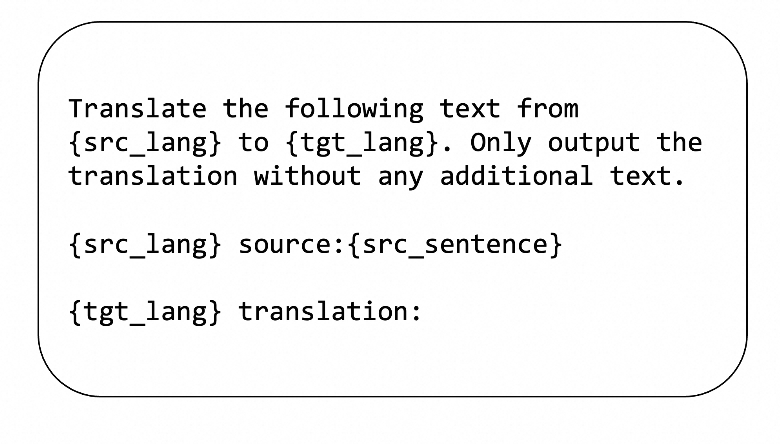}
  \caption{The prompt for flores evaluation}
  \label{fig:prompt_mt}
\end{figure}

The \textsc{Belebele} prompt is similar but includes a passage field for the context. Multilingual versions follow the same structure, translated accordingly.

\begin{figure*}
  \centering
  \includegraphics[width=1.0\linewidth]{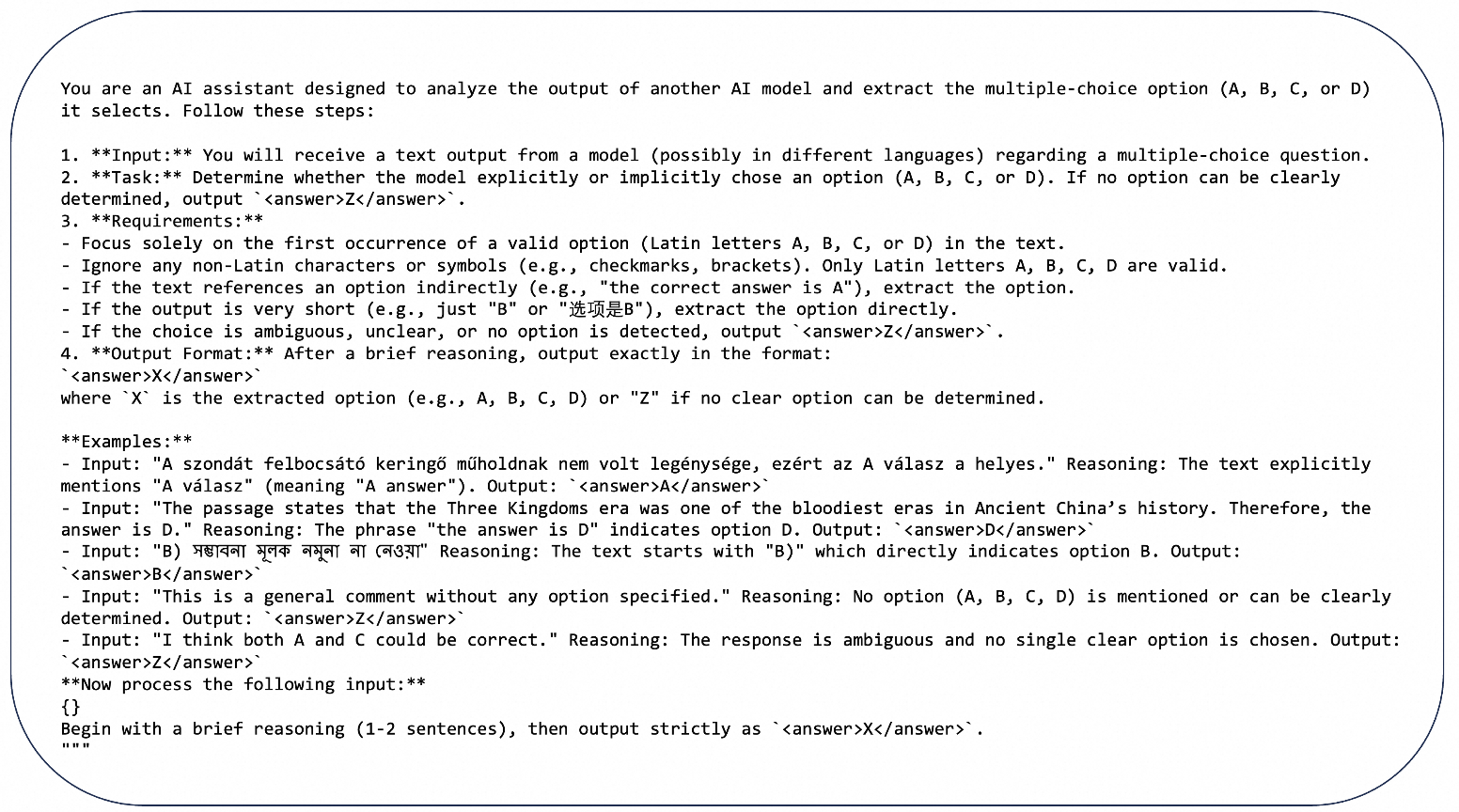}
  \caption{The prompt used for model-based answer extraction}
\label{fig:extraction_prompt}
\end{figure*}

\minisection{Robust Answer Extraction.}
We implement a two-stage process for robust answer extraction from model outputs.
\begin{enumerate}[leftmargin=*, itemsep=1pt, topsep=2pt]
    \item \textbf{Regex Extraction:} We first apply a regular expression to parse the final line of the model's generation, searching for the pattern `Answer: [A-D]`.
    \item \textbf{Model-based Extraction Fallback:} For outputs where regex parsing fails, we employ Qwen3-4B-Instruct\footnote{\url{https://huggingface.co/Qwen/Qwen3-4B-Instruct-2507}} as a fallback extractor. Prompted as detailed in Figure~\ref{fig:extraction_prompt}, it identifies the chosen option (A-D) or reports ambiguity ('Z'), ensuring robust answer parsing across varied response formats.
    \end{enumerate}

\section{Router-Tuning Replay Strategy}
\label{app:router_tuning}

To enhance knowledge retention for \method, we implemented a brief, router-tuning phase. In this stage, only the router parameters of the MoE layers were trained. The training data consisted of a 0.15B original languages token corpus, with a 1:2 ratio of original language data to expansion language data. This process allows the router to refine its ability to correctly allocate tokens from original languages to the frozen expert.

\section{Ablation on LLaMA-Pro Configuration}
\label{app:llamapro_ablation}

To ensure a fair and robust comparison against the LLaMA-Pro baseline, we conducted an ablation study to determine the optimal number of newly added Transformer blocks. We experimented with adding 7, 14, and 28 blocks to the base 28-layer Qwen2.5-7B architecture. This resulted in models with total parameter counts of 9.2B, 10.8B, and 12.4B, respectively.

As shown in Table~\ref{tab:llamapro_ablation_results}, the model with 14 added layers (10.8B total parameters) achieved the best overall performance, maximizing gains on expanded languages while maintaining strong performance on the original languages. 

\begin{table*}[h!]
\centering
\small
\setlength{\tabcolsep}{5pt}
\begin{tabular}{l c c c}
\toprule
\textbf{Model Configuration} & \thead{Total \\ Params (B)} & \thead{Avg. \\ (Expanded)} & \thead{Avg. \\ (Original)} \\
\midrule
\textit{Qwen2.5-7B-Instruct} & \textit{7.6} & \textit{59.83} & \textit{82.33} \\
\hdashline
LLaMA-Pro (+7 layers) & 9.2 & 66.81 & 75.86 \\
\textbf{LLaMA-Pro (+14 layers)} & \textbf{10.8} & \textbf{70.07} & \textbf{79.72} \\
LLaMA-Pro (+28 layers) & 12.4 & 24.60 & 8.66 \\
\bottomrule
\end{tabular}
\caption{Ablation on the number of added layers for the LLaMA-Pro baseline in Qwen2.5-7B. }
\label{tab:llamapro_ablation_results}
\end{table*}

\section{Ablation on LLaMA-Pro with Tulu Delta}
\label{app:llamapro_tulu_ablation}

Similar to our main experiments, we performed an ablation study to identify the strongest LLaMA-Pro configuration when using the Tulu alignment delta. We tested models with 7, 14, and 28 added layers.

The results, presented in Table~\ref{tab:llamapro_tulu_ablation_results}, show a different trend compared to our primary experiments. In this scenario, adding only 7 layers yielded the best performance. Consequently, we selected the LLaMA-Pro (+7 layers) variant as the baseline for our generalization analysis in Section 5.1.

\begin{table*}[h!]
\centering
\small
\setlength{\tabcolsep}{5pt}
\begin{tabular}{l c c c}
\toprule
\textbf{Model Configuration} & \thead{Total \\ Params (B)} & \thead{Avg. \\ (Expanded)} & \thead{Avg. \\ (Original)} \\
\midrule
\textit{Qwen-7B + Tulu-Delta} & \textit{7.6} & \textit{53.85} & \textit{79.90} \\
\hdashline
\textbf{LLaMA-Pro (+7 layers)} & \textbf{9.2} & \textbf{64.97} & \textbf{75.49} \\
LLaMA-Pro (+14 layers) & 10.8 & 32.71 & 38.96 \\
LLaMA-Pro (+28 layers) & 12.4 & 22.36 & 8.34 \\
\bottomrule
\end{tabular}
\caption{Ablation on the number of added layers for the LLaMA-Pro baseline using the Tulu delta. }
\label{tab:llamapro_tulu_ablation_results}
\end{table*}

\section{Ablation on LLaMA-Pro for LLaMA-3.1}
\label{app:llamapro_llama3_ablation}

To identify the strongest LLaMA-Pro baseline for the LLaMA-3.1-8B model family, we performed an ablation study on the number of added Transformer blocks. We tested variants with  8, 16, 32 added layers.

The results, presented in Table~\ref{tab:llamapro_llama3_ablation_results}, show a different trend compared to our primary experiments. In this scenario, adding only 8 layers yielded the best performance.

\begin{table*}[h!]
\centering
\small
\begin{tabular}{l c c c}
\toprule
\textbf{Model Configuration} & \thead{Total \\ Params (B)} & \thead{Avg. \\ (Expanded)} & \thead{Avg. \\ (Original)} \\
\midrule
\textit{LLaMA-3.1-8B-Instruct} & \textit{8.0} & \textit{64.91} & \textit{80.37} \\
\hdashline
\textbf{LLaMA-Pro~(+8 layers)} & 9.7 & \textbf{58.92} & 65.78 \\
LLaMA-Pro (+16 layers) & 11.5 & 44.44 & \textbf{66.49} \\
LLaMA-Pro (+24 layers) & 15 & 14.80 & 13.87 \\
\bottomrule
\end{tabular}
\caption{Ablation on the number of added layers for the LLaMA-Pro baseline in LLaMA3.1-8B.}
\label{tab:llamapro_llama3_ablation_results}
\end{table*}

\section{Expert Capacity and Language Scalability}
\label{app:expert_capacity}

In this section, we provides additional experiments to demonstrate that a small number of newly added trainable experts are sufficient to handle multiple expanded languages.

\paragraph{Expert count does not need to scale with language count.}
To investigate whether more experts are needed as additional languages are introduced, we fix the number of expanded languages to 5 (Hungarian, Bengali, Serbian, Telugu, and Icelandic, with 1B tokens each) and compare our standard 4-expert setup (1 frozen + 3 trainable) against a 6-expert setup (1 frozen + 5 trainable). As shown in Table~\ref{tab:expert_count_ablation}, increasing the number of experts yields negligible performance improvement, indicating that 3 trainable experts already provide sufficient capacity for this setting.

\begin{table*}[h!]
\centering
\small
\setlength{\tabcolsep}{5pt}
\begin{tabular}{l c c c c}
\toprule
\textbf{Model} & \textbf{Experts} & \thead{Original} & \thead{Expanded \\ (5-lang)} & \thead{Avg.} \\
\midrule
Qwen2.5-7B-Instruct & — & 82.33 & 51.81 & 67.07 \\
\hdashline
\method{} & 4 (1+3) & 81.26 & 70.07 & 75.67 \\
\method{} & 6 (1+5) & 81.03 & 70.66 & 75.84 \\
\bottomrule
\end{tabular}
\caption{Performance comparison when scaling the number of experts for 5 expanded languages. The marginal gain from increasing expert count confirms that 3 trainable experts are sufficient.}
\label{tab:expert_count_ablation}
\end{table*}

\paragraph{Scaling languages does not cause interference.}
We further examine whether expanding to more languages causes interference among learned languages. Keeping the expert count fixed at 4, we increased the number of expanded languages from three (Hungarian, Bengali, Serbian; totaling 9 billion tokens) to eight~(adding Czech, Telugu, Icelandic, Greek, and Turkish; 2 billion tokens each, amounting to 19 billion tokens in total). We evaluate on the original 3 expanded languages~(hu, bn, sr) to directly measure interference, and compare against the Dense-FT-Delta baseline trained on the same data.

As shown in Table~\ref{tab:language_scaling}, performance on the original 3 languages remains highly stable when scaling to 8 languages, and \method{} continues to substantially outperform the Dense-FT-Delta baseline. These results demonstrate that the trainable experts can accommodate an increasing number of languages without causing significant interference.

\begin{table*}[h!]
\centering
\small
\setlength{\tabcolsep}{4pt}
\begin{tabular}{l c c c c}
\toprule
\textbf{Model} & \textbf{Training Setup} & \thead{Original} & \thead{Expanded \\ (Initial 3)} & \thead{Avg.} \\
\midrule
Qwen2.5-7B-Instruct & — & 82.33 & 59.83 & 71.08 \\
\hdashline
\method{} & 3 langs (9B) & 81.17 & 74.97 & 78.07 \\
\method{} & 8 langs (19B) & 81.04 & 74.43 & 77.74 \\
Dense-FT-Delta & 8 langs (19B) & 78.21 & 70.00 & 74.10 \\
\bottomrule
\end{tabular}
\caption{Performance on the original 3 expanded languages when scaling to 8 languages with a fixed 4-expert setup. \method{} retains strong performance on previously learned languages and outperforms Dense-FT-Delta under identical data conditions.}
\label{tab:language_scaling}
\end{table*}

\end{document}